\documentclass[final,3p,times,twocolumn]{elsarticle}

\usepackage{amssymb}
\usepackage{amsmath}
\usepackage{amsfonts}
\usepackage{graphicx}

\usepackage{balance}       % to better equalize the last page
\usepackage{graphicx}
\usepackage[T1]{fontenc}   % for umlauts and other diaeresis
\usepackage{txfonts}
\usepackage[pdflang={en-US},pdftex]{hyperref}
\usepackage{color}
\usepackage{booktabs}
\usepackage{textcomp}
\usepackage{algorithm} %format of the algorithm 
\usepackage{algorithmic} %format of the algorithm 
\usepackage{multirow} %multirow for format of table 
\usepackage{xcolor}

\usepackage{microtype}        % Improved Tracking and Kerning
\usepackage{ccicons}     

\usepackage{lineno}

\newcommand{\re}[1]{\textcolor{black}{#1}}

\journal{arXiv}

\begin{document}

\begin{frontmatter}



\title{Riemannian Manifold Optimization for Discriminant Subspace Learning}


\author[1]{Wanguang Yin%
	}
\ead{yinwg@sustech.edu.cn}
\author[2]{Zhengming Ma
    }
\ead{issmzm@mail.sysu.edu.cn}
\author[1]{Quanying Liu\corref{cor1}}
\ead{liuqy@sustech.edu.cn}

\cortext[cor1]{Corresponding author}
\address[1]{Shenzhen Key Laboratory of Smart Healthcare Engineering, Department of Biomedical and Information Engineering, Southern University of Science and Technology, Shenzhen, 518055 China}
\address[2]{School of Electronics and Information Technology, Sun Yat-sen University, Guangzhou, Guangdong, 510006 China}

\begin{abstract}
Discriminant analysis, as a widely used approach in machine learning to extract low-dimensional features from the high-dimensional data, applies the Fisher discriminant criterion to find the orthogonal discriminant projection subspace. But most of the Euclidean-based algorithms for discriminant analysis are easily convergent to a spurious local minima and hardly obtain an unique solution. To address such problem, in this study we propose a novel method named Riemannian-based Discriminant Analysis (RDA), which transforms the traditional Euclidean-based methods to the Riemannian manifold space. In RDA, the second-order geometry of trust-region methods is utilized to learn the discriminant bases. To validate the efficiency and effectiveness of RDA, we conduct a variety of experiments on image classification tasks. The numerical results suggest that RDA can effectively extract low-dimensional features and robustly outperform state-of-the-art algorithms in classification tasks. 
\end{abstract}

\begin{keyword}
Dimensionality reduction \sep Discriminant analysis \sep Riemannian manifold optimization \sep Stiefel manifold \sep Grassmannian manifold

\end{keyword}

\end{frontmatter}

\section{Introduction}
Linear discriminant analysis (LDA) is an essential method for extracting statistically significant features as a prerequisite for pattern recognition and machine learning. LDA has broad applications ranging from text mining~\cite{radovanovic2008text} and image classification~\cite{savas2007handwritten} to brain-computer interface (BCI)~\cite{aldea2014classifications}. Generally, LDA learns to discriminate different classes by computing the distance (or similarity) metrics among the extracted features from training data, and then assign the test data to a specific class based on the measured distance and the learned threshold.  Therefore, the performance of LDA largely relies on the distance metrics defined on the features and the optimization strategy for solving the loss function. However, most current methods for solving LDA are based on the Euclidean space. However, these Euclidean-based methods easily convergent to a spurious local minimum and hardly obtain a globally optimal solution~\cite{chen2018solving}. \re{It motivates us to pursue alternative methods for solving the LDA and ensuring an effective approximation of the high-dimensional input data with a lower-dimensional representation.} 

To this end, by employing the specific nature of orthogonal constraints of the discriminant bases, the LDA can be transformed from the Euclidean space to a Riemannian manifold space and be solved by Riemannian manifold optimization. Specifically, the Riemannian manifold optimization utilizes underlying structures of the matrix manifold and optimizes the loss function by using the Riemannian-based conjugate gradient and trust-region method, which benefits from the Riemannian concepts, such as the tangent space, Riemannian metrics, Retraction, connection, and transport parallel \cite{bouchard2018riemannian}. It's worth noting that the trust-region method can linearly approximate a local solution on the tangent space in each iteration and eventually converge upon an extreme point as the globally nonlinear solution, usually resulting in superior performance compared to the traditional Euclidean-based methods \cite{bouchard2018riemannian}.

In this way, we propose a family of discriminant analysis algorithms defined on the Riemannian space, namely Riemannian-based discriminant analysis (RDA) (\re{\textbf{Sec} \ref{method}}). The performance of RDA is compared with Euclidean-based methods and several other existing Riemannian-based methods in terms of dimensionality reduction and classification. Our results show that RDA algorithms are superior in solving the multiclass, large-scale clustering tasks, as well as the classification tasks, compared to the Euclidean-based discriminant analysis. The main contributions of this paper can be concluded as the following:

\begin{itemize}
  \item First, RDA transforms the linear discriminant analysis from the Euclidean space to the Riemannian manifold space and then employs the trust-region method to learn the discriminant basis of the projection subspace (\re{\textbf{Sec} \ref{method}}). In this way, the loss function can be converted from a division form to a subtraction one (\re{Eq.(\ref{cost10})}). RDA can therefore effectively avoid calculating the inverse of the Hessian matrix.
  \item Second, two types of Riemannian manifolds (\textit{i.e.} Stiefel manifold and Grassmann manifold) are investigated, and effects of the second-order approximation and the sparsity regularization on the discriminant bases are constructed. The numerical experiments suggest that the second-order geometry of the trust-region method on the Riemannian manifold outperforms the first-order geometry of the conjugate gradient method. 
  %\item Third, in order to reduce the learning parameters, we adopt a sparsity regularization term in terms of discriminant bases to investigate the generalization ability of our proposed model.
  \item Lastly, RDA achieves state-of-the-art (SOTA) performance in both clustering experiments (\re{\textbf{Sec} \ref{exp:clustering}}) and classification experiments (\re{\textbf{Sec} \ref{exp:classification}}). The numerical experiments on multiple image datasets (\textit{e.g.} COIL20, ETH80, MNIST, USPS, CMU PIE) demonstrate that RDA can robustly obtain higher performance than traditional Euclidean-based algorithms, as well as other existing Riemannian-based algorithms.
\end{itemize}

% The rest of the paper is structured as follows. In section 2, we review the related work on matrix and tensor decomposition for subspace learning, as well as the Riemannian manifold optimization. In section 3, we propose RDA and present a detailed description of the solving process, including Riemannian gradient, Riemannian Hessian, and sparsity regularization. Section 4 conducts numerical experiments on seven image datasets and compares RDA with other existing algorithms. Finally, in section 5, we conclude this paper and prospective to our future work.

 \section{Related Work}
\subsection{Subspace Learning}
\label{sec2.1}
Subspace learning is essential for computer vision, pattern recognition \cite{lu2003face}, biomedical engineering \cite{li2008prior, lu2009regularized}, and bioinformatics \cite{ye2004using}. It aims to map the high-dimensional data to a lower-dimensional space with maximally maintaining the information in the original data. The input data is usually represented as vectors, matrices, or tensors, and subspace learning is to find an optimal mapping, either linear or nonlinear, to project the input data to a low-dimensional space. Linear subspace learning is a powerful tool for dimensionality reduction and it provides a solid foundation for machine learning algorithms~\cite{jiang2011linear}. A variety of methods have been proposed for linear subspace learning, including the matrix or tensor decomposition~\cite{phan2010tensor, nie2010flexible} and the linear discriminant analysis~\cite{wang2019unsupervised}. 

The representative algorithms for matrix factorization include singular value decomposition (SVD) \cite{paige1981towards}, principle component ananlysis (PCA) \cite{wold1987principal}, canonical correlation analysis (CCA) \cite{johnson2002applied}, independent component analysis (ICA) \cite{comon1994independent}, and nonnegative matrix factorization (NMF) \cite{lee2001algorithms}. 
%Specifically, SVD is a factorization of a matrix that generalizes the eigendecomposition of the square normal matrix. 
%PCA maximizes the mutual information between the original high-dimensional Gaussian distributed data and projected low-dimensional space. CCA measures the correlation coefficient between two variables, as an extension of the Pearson correlation to the multivariate context \cite{helmer2020stability}. 
%ICA decomposes the variable matrix as statistically independent as possible, which is widely applied in blind source separation \cite{bouchard2018riemannian}. 
NMF assumes that the original input data is nonnegative, and the components as a part-based representation of the original data are also nonnegative. 
Moreover, tensor decomposition, as an extension of matrix factorization to the higher-order arrays, is ubiquitously used for linear subspace learning for high-order data.
Tucker decomposition and the canonical decomposition are two main types of tensor decomposition methods. The former is usually used in machine learning, and the latter is usually used in signal processing, also known as the parallel factors (PARAFAC) decomposition. Higher-order orthogonal iteration (HOOI) is a variant of Tucker decomposition with orthogonality constraints in the projection matrices \cite{de2000best}.
The higher-order singular value decomposition (HOSVD) extends the matrix SVD to higher-order tensors, and its projection matrices are column-wise orthogonality and the core tensor is orthogonal as well. Its computation leads to the calculation of $ N $ different matrix SVDs of differently unfolded matrices \cite{de2000multilinear, lu2011survey}. Multilinear CCA, as a multilinear extension of the CCA algorithm, aims to find maximal correlations between the weighted linear combinations of variables \cite{vasilescu2005multilinear}. Multilinear PCA aims to find a tensor to tensor projection that maximally captures the variations of the original tensorial data \cite{lu2008mpca} . Multilinear ICA model of tensor data learns the statistically independent component of multiple factors \cite{vasilescu2005multilinear}. When the components of the raw input data is nonnegative, especially when it meets to the nonnegative conditions have the physical meaning, such as spectrum, energy, and probability, it hence that nonnegative tensor factorization (NTF) is enforced nonnegative conditions on the PARAFAC model to find the nonnegative factors or components \cite{hazan2005sparse}. \re{Non-negative Tucker Decomposition (NTD) is based on Tucker tensor decomposition and simultaneously enforces non-negative constraints on the projection matrix and the core tensor\cite{kim2007nonnegative}. Low-rank regularized heterogeneous tensor decomposition (LRRHTD) adds the orthogonal constraint for the first N-1 modes and the low-rank constraint for the last mode of the projection matrix \cite{zhang2017low}.}

Alternatively, LDA and its variants are another popular way for subspace learning, especially when the labelled data is available \cite{fisher1936use}. The target of LDA is to find a discriminant subspace that maximizes the trace of the between-class scatter while minimizing the trace of the within-class scatter. Some variants of LDA have been proposed~\cite{sifaou2020high,su2019order}, including the discriminant analysis with tensor representation (DATER) ~\cite{yan2005discriminant} and the general tensor discriminant analysis (GTDA) \cite{tao2007general}. The discriminant analysis with tensor representation (DATER) algorithm aims to find a tensor-to-tensor projection while maximizing the tensor-based scatter ratio \cite{yan2005discriminant}. However, a limitation of this algorithm is that it does not always converge over its iterations. The general tensor discriminant analysis (GTDA) learns a discriminant subspace with a tensor-to-tensor projection while maximizing the discriminant information in a low-dimensional space \cite{li2014multilinear}. Consider that independence between extracted features is a desirable property in many real-world applications, such that, uncorrelated multilinear discriminant analysis (UMLDA) has been proposed to extract uncorrelated discriminative features directly from tensorial data, with an assumption that each class is represented by a single cluster and none of them can be solved by nonlinear separation~\cite{lu2008uncorrelated}. Moreover, the tensor rank-one discriminant analysis (TR1DA) is to learn the projection subspace by repeatedly calculating the residues of the original data with the scatter difference criterion, and eventually obtains a set of rank-one projections~\cite{tao2008tensor}. The high order discriminant analysis (HODA) is to find discriminative bases that based on the multilinear structure of Tucker model~\cite{phan2010tensor}. \re{The constrained multilinear discriminant analysis (CMDA) seeks an optimal tensor-to-tensor projection for discrimination in a lower-dimensional tensor subspace \cite{li2014multilinear}. Theoretically, the value of the scatter ratio criterion in CMDA approaches its extreme value, if it is exists, with a bounded error.}
    
Although the methods for linear subspace learning are well-studied, there are still a number of open challenges, regarding the effectiveness and the robustness in characterizing the nonlinear structures of the high-dimensional data. In fact, several studies have reported that DATER could not guarantee convergence to a stationary point during iterations~\cite{phan2010tensor, li2014multilinear}. 
Another critical issue of LDA-type algorithms are the singularity and instability of the within-class scatter. 
%Robustness implies that they are reliably estimated across different data sets from the same population, despite inherent variability in the data. In addition, most of aforementioned algorithms are to learn the discriminant bases by maximizing the between-class scatter and minimizing the within-class scatter in Euclidean space. In the situation that tensor data to be vectorized, resulting in the observation vectors are longer than the number of observations. 
Since LDA and its variants rely on the calculation of the discriminant score, while the discriminant score requires computing the inverse of the covariance matrix \cite{sifaou2020high}, thus it might meets the singularity problem. To address such problems, Riemannian manifold optimization is considered an candidate approach to learn the discriminant projection subspace.

\subsection{Riemannian Manifold Optimization}
\label{sec2.2}
\begin{figure}[htb]
      \centering
      \includegraphics[width=1\linewidth]{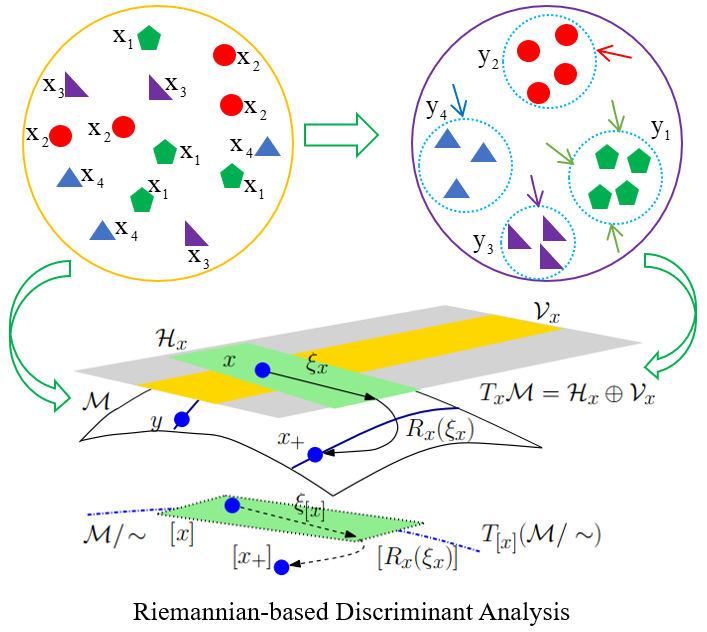}
      \caption{A semantic illustration of the Riemannian-based Discriminant Analysis. Here, $\mathrm{x}$ and $\mathrm{y}$ respectively represent the raw input data and the reduced output data in the Euclidean space. For the Riemannnian manifold optimization, $\mathcal{M}$ denotes the matrix manifold, and its tangent space $T_x\mathcal{M}$ is a tangent plane on a point $x$ of the manifold $\mathcal{M}$, which can be divided into the horizontal space $ \mathcal{H}_x$ and the vertical space $\mathcal{V}_x$. A retraction $ R_x\left ( \xi _x \right ) $ is a mapping from the tangent space back onto the manifold $\mathcal{M}$. The tangent vector $ \xi_x $ on the tangent space $ T_x\mathcal{M} $ denotes a possible movement direction at point $ x $.}
      \label{fig:Optimization}
\end{figure}

Riemannian manifold is actually a smooth subset of a vector space included in the Euclidean space \cite{cruceru2020computationally}. It abandons the flat Euclidean space and formulates the optimization problem directly on the curved manifold. To describe a general framework of Riemannian manifold optimization, it needs to define some basic ingredients, such as the Riemannian matrix manifold $ \mathcal{M} $, smooth function $ f:\mathcal{M}\rightarrow \mathbb{R} $ (\textit{i.e.} along with its Riemannian gradient \textit{i.e.} $ \mathrm{grad}f $, or Riemannian Hessian \textit{i.e.} $ \mathrm{hess}f $ to perform the procedures of Riemannian manifold optimization), projection operator \textit{i.e.} $ \mathrm{P}^t\left ( \cdot  \right ) $, Riemannian metric \textit{i.e.} $ \mathrm{g}\left ( \cdot ,\cdot  \right ) $, Riemannian connection \textit{i.e.} $ \bigtriangledown_\xi \eta  $, and retraction \textit{i.e.} $ R_x\left ( \xi  \right ) $. Concretely, we can define a \textit{projection operator} to project the embedded space (\textit{i.e.} ambient space) to its tangent space, that is obtained by subtracting the component in the orthogonal complement of the tangent space (\textit{i.e.} normal space $ \mathcal{N}_x $). If the Riemannian manifold is a quotient manifold, we can further define a \textit{projection operator} from the tangent space to the horizontal space, that is obtained by removing the component in the orthogonal complement of the horizontal space (\textit{i.e.} vertical space $ \mathcal{V}_x $). Note that \textit{connection} is an important notion that intimately relevant to the Riemannian Hessian and the \textit{vector transport}, and Levi-Civita connection is a unique affine connection used to define the Riemannian Hessian of a loss function \cite{absil2009optimization}. \textit{Vector transport} allows movements from a tangent space to another tangent space. \textit{Retraction} is a mapping from the tangent space back onto the manifold, ensuring that each update of Riemannian manifold optimization remains on the manifold, and the exponential retraction is the most expensive retraction, which describes the movement along a \textit{geodesic}. A \textit{geodesic} is defined as a curve with the minimal length connecting two points on the manifold. 
\textbf{Figure~\ref{fig:Optimization}} is the semantic illustration of the Riemannian-based discriminant analysis. $ T_x\mathcal{M} $ is the tangent space of the embedded matrix manifold $ \mathcal{M} $ endowed with a bilinear, symmetric-positive form of Riemannian metric \textit{i.e.} $ \mathrm{g}\left ( \cdot ,\cdot  \right ) $, that is termed as a Riemannian manifold. In other words, a Riemannian manifold is a smooth manifold with a Riemannian metirc. The Riemannian metirc defines a family of inner products on the tangent spaces that smoothly vary with point $ x $ on the manifold. Once that Riemannian metric is defined, the distance, angle, and the curvature on the manifold can be calculated.

The Stiefel manifold and the Grassmann manifold are two popular manifolds to conduct Riemannian manifold optimization. Specifically, the \textit{Stiefel manifold} $ \mathrm{St}\left ( D,d \right ) $ is a set of $ D\times d $ orthonormal matrices $ \left \{ U\in\mathbb{R}^{D\times d}:U^TU=I_d \right \} $~\cite{absil2009optimization}. Notably, Stiefel manifold has no unique representation of $U$, for multiplying by any orthogonal identity group does not change its original representation. Thus, if $ O_d $ is a set of $ d\times d $ orthogonal matrices, then $ U_1=U_2O_d $. Otherwise, the \textit{Grassmann manifold} $ \mathrm{Gr}\left ( D,d \right ) $ is a set of $ d $-dimensional linear subspace of $ \mathbb{R}^D $ \cite{absil2009optimization,de2016nonlinearly}. If $ d\leq D $, then the elements on the Grassmann manifold $ U\in\mathrm{Gr}\left ( D, d \right ) $ can also be represented as the column space of Stiefel manifold $ U\in\mathrm{St}\left ( D,d \right ) $, that is identified with a set of equivalent classes $ \left [ U \right ]\in\mathrm{Gr}\left ( D,d \right ) $. Additionally, some notions closely relevant to the Riemannian manifold (\textit{e.g.} the Riemannian metric, tangent space, and tangent vector) are worthy to clarify. When the columns of equivalence class $ \left [ U \right ] $ equals to the columns of $ U $, such as for given $ U\in\mathrm{St}\left ( D,d \right ) $, the inner product of $ \mathrm{St}\left ( D,d \right ) $ also holds for $ \mathrm{Gr}\left ( D,d \right ) $, and the \textit{tangent space} $ T_U\mathrm{St}\left ( D,d \right ) $ of Stiefel manifold is a vector space of all tangent vectors at point $ U $, and the \textit{tangent vector} $ \xi $ on the tangent space $ T_U\mathrm{St}\left ( D,d \right ) $ is a possible movement direction at point $ U $, characterized as a matrix of $ D\times d $. 

\re{Many existing methods for subspace learning in Sec \ref{sec2.2} can be extended to Riemannian manifold space. For example, manifold-based high order discriminant analysis (MHODA) is an extension of HODA from Euclidean space to the Riemannian manifold space ~\cite{yin2020high}. Taking into account of the heterogeneity in multimodal data, HTD Multinomial add the orthogonal constraints on the first $N-1$ modes of the corresponding projection matrices, while the last mode of the corresponding sample information is treated as the Multinomial manifold. It results in an optimization problem on the Multinomial manifold which can be solved by using the second-order geometry of trust-region method \cite{sun2015heterogeneous}.}

\re{Naturally, using Riemannian manifold optimization can uncover the nonlinear geometric structures of the high-dimensional data. It has valuable merits to guarantee convergence to a globally optimal solution, whereas the traditional methods (\textit{e.g.} alternating least square (ALS) \cite{holtz2012alternating}, multiplicative updating rules (MURs) \cite{kim2007nonnegative}, and alternating direction method of multipliers (ADMM) \cite{boyd2011distributed}) might be stuck into the local minima. In this work, we propose a novel RDA method, which performs Riemannian manifold optimization for discriminant subspace learning. Specifically, we define the loss function of RDA in the Riemannian space, derive the Riemannian Hessian and present Riemannian optimization algorithms for RDA.} 

\section{Riemannian-based Discriminant Analysis (RDA)}
\label{method}

\subsection{The Loss Function of RDA}
The target of linear discriminant analysis is to minimize the reconstruction error in a mapping from high-dimensional data to a low-dimensional feature space, while maximizing the discrimination between classes. In other words, it aims to find an optimal discriminant bases $ U\in\mathbb{R}^{D\times d} $ by minimizing the within-class scatter $ S_W $ and maximizing the between-class scatter $ S_B $, whereas the manipulation of projection operation $ \mathrm{y}=U^T\mathrm{x} $, and matrix $U$ is subject to the orthogonal constraint \textit{i.e.} $ U^T U = I_d $. Here, we denote $ \mathrm{x}\in\mathbb{R}^D $ as the input data with a high dimension $D$, $ \mathrm{y}\in\mathbb{R}^d $ as the low-dimensional representation of the input data. More concretely, the loss function $f\left ( U \right )$ can be formulated as the following:

%\begin{equation}\label{cost}
%\mathop{\mathrm{min}}_U\left ( S_W - S_B \right )=\sum_{c=1}^{C}\sum_{n\in C_c}\left \| \mathrm{y}_n - \overline{\mathrm{y}}_{c} \right \|_F^2 - \sum_{c=1}^{C}N_c\left \| \overline{\mathrm{y}}_c - \overline{\mathrm{y}} \right \|_F^2
%\end{equation}
\begin{equation}\label{cost}
\begin{aligned}
& \mathop{\mathrm{min}}_U f\left ( U \right )\\
& = \sum_{c=1}^C\sum_{n\in C_c}\left \| \mathrm{y}_n-\overline{\mathrm{y}}_{c} \right \|_F^2-\sum_{c=1}^CN_c\left \| \overline{\mathrm{y}}_c - \overline{\mathrm{y}} \right \|_F^2\\
& = \sum_{c=1}^C\sum_{n\in C_c}\left \| U^T\left ( \mathrm{x}_n-\overline{\mathrm{x}}_{c} \right ) \right \|_F^2- \sum_{c=1}^CN_c\left \| U^T\left ( \overline{\mathrm{x}}_c - \overline{\mathrm{x}} \right ) \right \|_F^2\\
& = \left \| U^T\left ( X-\overline{X}_{C} \right ) \right \|_F^2 - \left \| U^T\left ( \overline{X}_C - \overline{X} \right ) \right \|_F^2\\
& = tr\left ( U^TS_WU \right ) - tr\left ( U^TS_BU \right )\\
& \mathrm{s.t. } \   U^T U = I_d
\end{aligned}
\end{equation}
where $ N $ is the number of samples and $ N_c $ is the number of samples from class $ c $. Obviously, the number of samples $ N=\sum_{c=1}^CN_c $, and the sample mean $ \overline{\mathrm{y}}=\frac{1}{N}\sum_n\mathrm{y}_n $. The mean of samples from class $ c $, denoted as $\overline{\mathrm{y}}_c$, with $\overline{\mathrm{y}}_c=\frac{1}{n_c}\sum_n\left [ \mathrm{y}_n|n=c \right ] $.  
$ S_W=\left ( X-\overline{X}_{C} \right )\left ( X-\overline{X}_{C} \right )^T $ is a covariance matrix relative to the within-class scatter, and $ S_B=\left ( \overline{X}_C-\overline{X} \right )\left ( \overline{X}_C-\overline{X} \right )^T $ is a covariance matrix relative to the between-class scatter. Note that the procedures of categorical alignment can promote transferable learning and strengthen the generalization ability of the model. 

An advantage of the loss function in Eq.(\ref{cost}) is that we convert the divisive form to the subtractive one, thereby allows to effectively calculate the Riemann gradient and Riemann Hessian. We can transform the constrained loss function of Eq.(\ref{cost}) in Euclidean space to an unconstrained one in Stiefel manifold $ U\in{\mathrm{St}}\left ( D,d \right ) $, and then employ Riemannian manifold optimization to solve the loss function. Therefore, the loss function in the Stiefel manifold can be rewritten as: 

\begin{equation}\label{cost1}
\begin{aligned}
\mathop{\mathrm{min}}_{U\in{\mathrm{St}}(D,d)}f\left ( U \right ) 
%&=\sum_{c=1}^C\sum_{n\in C_c}\left \| \mathrm{y}_n-\overline{\mathrm{y}}_{c} \right \|_F^2-\sum_{c=1}^CN_c\left \| \overline{\mathrm{y}}_c - \overline{\mathrm{y}} \right \|_F^2\\
%&= \sum_{c=1}^C\sum_{n\in C_c}\left \| U^T\left ( \mathrm{x}_n-\overline{\mathrm{x}}_{c} \right ) \right \|_F^2- \sum_{c=1}^CN_c\left \| U^T\left ( \overline{\mathrm{x}}_c - \overline{\mathrm{x}} \right ) \right \|_F^2\\
%=\left \| U^T\left ( X-\overline{X}_{C} \right ) \right \|_F^2 - \left \| U^T\left ( \overline{X}_C - \overline{X} \right ) \right \|_F^2\\
=tr\left ( U^TS_WU \right ) - tr\left ( U^TS_BU \right )\\
\end{aligned}
\end{equation}

\re{According to} the equivalence relation defined by the orthogonal group $ \mathcal{O}\left( d \right)$, the Grassmann manifold $ \mathrm{Gr}\left ( D,d \right ) $ can be formulated as the quotient space of Stiefel manifold. In this case, the loss function of Eq.(\ref{cost}) can be formulated on the Grassmann manifold:

\begin{equation}\label{cost2}
\mathop{\mathrm{min}}_{[U]\in{\mathrm{Gr}}(D,d)}f\left ( U \right ) =tr\left ( U^TS_WU \right ) - tr\left ( U^TS_BU \right )
\end{equation}
where $ \left [ U \right ]\in{\mathrm{Gr}}\left ( D,d \right ) $ is the equivalence class for a given $ U\in\mathrm{St}\left ( D,d \right ) $, and $ \left [ U \right ] $ denotes a Grassmann point. 

Since the covariance matrix is a symmetric-positive definite matrix, then the optimization problem of Eq.(\ref{cost1}) can also be formulated on the generalized Stiefel manifold $\mathrm{GSt}\left ( D,d;G \right )$ as

\begin{equation}\label{cost3}
\mathrm{GSt}\left ( D,d;G \right )=\left \{ U\in\mathbb{R}^{D\times d}:U^TGU=I_d \right \}
\end{equation}
where $ G $ denotes a covariance matrix. 

Similarly, the optimization problem of Eq.(\ref{cost2}) can be cast on the generalized Grassmann manifold $\mathrm{GGr}\left ( D,d;G \right )$ as

\begin{equation}\label{cost4}
\mathrm{GGr}\left ( D,d;G \right )=\mathrm{GSt}\left ( D,d;G \right )/\mathcal{O}\left ( d \right )
\end{equation}
where $ \mathcal{O}\left( d \right)$ represents the orthogonal group.

\subsection{The Learning Algorithm for RDA}
Here we present some of typical objects relative to the embedded submanifold that utilized in the Riemannian manifold optimization. Firstly, the loss function of Eq.(\ref{cost2}) is reformulated as following:

\begin{equation}\label{cost6}
f\left ( U \right )=tr\left ( U^T\left ( S_W - S_B \right )U \right )
\end{equation}

We use the inner product $ \mathrm{g_U}:T_U\mathcal{M}\times T_U\mathcal{M}\rightarrow \mathbb{R} $ as the Riemannian metric on the tangent space of the manifold: 

\begin{equation}\label{cost7}
\mathrm{g}_U\left ( \xi ,\eta  \right )=tr\left ( \xi ^T\eta \right )
\end{equation}

In addition, the covariance matrix $ G $ can be defined as the scaling matrix of Riemannian preconditioning that regulates Riemannian metric on the tangent space:

\begin{equation}\label{cost8}
\mathrm{g}_U\left ( \xi ,\eta  \right )=tr\left ( \xi ^T\eta/G \right )
\end{equation}

We denote $\nabla f$ as the Euclidean gradient of the loss function Eq.(\ref{cost6}), and obtain the following expression:

\begin{equation}\label{cost9}
\nabla f\left ( U \right )=2 S_W U - 2 S_B U
\end{equation}

Once the computational space is split into two complementary spaces (\textit {i.e.} the tangent space and normal space), the Riemannian gradient of loss function, denoted as $ \mathrm{grad}f\left ( U \right ) $, can be obtained by the orthogonal projection of the Euclidean gradient $ \nabla f\left ( U \right ) $ to the tangent space of the Riemannian manifold. For the Stiefel manifold, $ \mathrm{grad}f(U) $ can be calculated as follows

\begin{equation}\label{cost10}
\begin{aligned}
\mathrm{grad}f\left ( U \right )&=\mathrm{P}_U^t\left ( \nabla f\left ( U \right ) \right )\\
& = \nabla f\left ( U \right ) - U\mathrm{sym}\left ( U^T\nabla f\left ( U \right ) \right )\\
\end{aligned}
\end{equation}
\re{where the function $ \mathrm{sym}\left ( X \right )$ is defined as $ \mathrm{sym}\left ( X \right )=\left ( X+X^T \right )/2 $ to extract the symmetric part of a square matrix $ X $}. 

In the case that generalized Stiefel manifold, its orthogonal projection of Euclidean gradient $ \nabla f\left ( U \right ) $ from an ambient space to the tangent space can be efficiently computed by the following

\begin{equation}\label{cost11}
\begin{aligned}
&\mathrm{grad}f\left ( U \right )\\
&=\mathrm{P}_U^t\left ( \nabla f\left ( U \right ) \right )\\
&=\nabla f\left ( U \right ) - U\mathrm{sym}\left ( U^TG \nabla f \left ( U \right ) \right )\\
\end{aligned}
\end{equation}

\re{Likewise}, the orthogonal projection from an ambient space to the tangent space for the generalized Grassmann manifold can be formulated as following

\begin{equation}\label{cost12}
\mathrm{P}_{\left [ U \right ]}^t\left ( U \right )=U-U\mathrm{sym}\left ( U^TGU \right )
\end{equation}

Note that the second-order geometry of Riemannian Hessian is one of the most important concepts relative to the connection \textit{i.e.} $ \bigtriangledown _\xi \eta $, denoting the covariant derivative of the vector field $ \eta $ along the direction of another vector field $ \xi $. Given a concrete example, the covariant derivative of $ D\nabla f\left ( U \right )\left [ \xi  \right ] $ is the Euclidean directional derivative of the Euclidean gradient $ \nabla f\left ( U \right ) $ along the direction of the tangent vector $ \xi $ on the manifold. 

Therefore, the Euclidean Hessian, \textit{i.e.} $ \mathrm{Hess}f\left ( U \right )\left [ \xi  \right ] $, can be directly calculated from Eq.(\ref{cost9}) as following:

\begin{equation}\label{cost13}
\begin{aligned}
\mathrm{Hess}f\left ( U \right )\left [ \xi  \right ]&=D \nabla f \left ( U \right )\left [ \xi  \right ]\\
&=2S_W\xi -2S_B\xi \\
\end{aligned}
\end{equation}

And, the Riemannian Hessian, \textit{i.e.} $ \mathrm{hess}f\left ( U \right )\left [ \xi  \right ] $, equals to the Euclidean Hessian followed by the orthogonal projection onto the tangent space equipped with Riemannian metric, thus

\begin{equation}\label{cost14}
\mathrm{hess}f\left ( U \right )\left [ \xi  \right ]=\mathrm{P}_U^t\left ( \mathrm{Hess}f\left ( U \right )\left [ \xi  \right ] \right )
\end{equation}

For the Riemannian quotient manifold (\textit{e.g.} Grassmann manifold), it requires to further split the tangent space into other two orthogonal complementary subspaces (\textit {i.e.} the horizontal space and vertical space). Then, we can conduct an orthogonal projection from the tangent space to the horizontal space along the equivalence class of the vertical space to effectively isolate the extreme point as the unique solution. More detailed discussions about the quotient space refer to~\cite{absil2009optimization,mishra2016riemannian}. For the implementation of Riemannian manifold optimization, Riemannian version of conjugate gradient, and steepest descent, and trust-region method have been constructed into some existing toolbox, such as the Manopt toolbox \cite{boumal2014manopt}. Once the Riemannian gradient in Eq.(\ref{cost10}) and Riemannian Hessian in Eq.(\ref{cost14}) are calculated, it is convenient to perform Riemannian manifold optimization for solving the RDA.

\subsection{Sparsity regularized discriminant analysis}
In this subsection, we take into account of the model's generalization ability, and further incorporate an additional term about $ U $ into the loss function to prevent the model from overfitting the data. Specifically, the loss function $f(U)$ for the sparsity regularized discriminant analysis is formulated as follows: 

\begin{equation}\label{cost15}
\mathop{\mathrm{min}}_{U}f(U) = tr\left ( U^T \left(S_W - S_B \right) U \right ) + \lambda \left \| U \right \|_1
\end{equation}
where $ \lambda $ is a hyper-parameter to balance the discriminant performance and the sparsity of $U$ in the model. Here, the loss function $f(U)$ can be defined on either Stiefel manifold or Grassmann manifold. 
%Imposing the constraint on $ U $ to achieve more robust estimation of parameters, where $ \left \| U \right \|_1 $ is the sum of the absolute values of the entries of the matrix $ U $. 

To solve the sparsity regularized discriminant analysis, we have to derive the first-order and second-order derivatives of regularization term \textit{w.r.t.} $ U $. Naturally, we can obtain the Euclidean gradient of $\left \| U \right \|_1$ \textit{w.r.t.} $ U $ as $ \nabla \left \| U \right \|_1=\mathrm{sgn}\left ( U \right ) $, where

\begin{equation}\label{cost16}
\mathrm{sgn}\left ( U \right ) = \left\{\begin{matrix}
1  & if & \mathrm{U}\left ( i,j \right )> 0\\ 
0  & if & \mathrm{U}\left ( i,j \right )= 0\\ 
-1 & if & \mathrm{U}\left ( i,j \right )< 0
\end{matrix}\right.
\end{equation}

And, the second-order derivatives of $\left \| U \right \|_1$ with respect to $ U $ in the Euclidean space is obtained as follows:

\begin{equation}\label{cost17}
\mathrm{Hess}\left \| U \right \|_1 = 2\sigma \left ( U \right )
\end{equation}

where $ \sigma \left ( U \right ) $ is defined as:

\begin{equation}\label{cost18}
\sigma \left ( U \right )=\left\{\begin{matrix}
1 & if \; \; U\left ( i,j \right )=0\\ 
0 & otherwise
\end{matrix}\right.
\end{equation}

Till now, all the deductions about loss function of the sparsity regularized discriminant analysis have been completed. Algorithm 1 provides the pseudo code of the optimization procedures. The code for RDA is available at \url{https://github.com/ncclabsustech/RDA-algorithm}.

\begin{algorithm} %算法开始 
	\caption{Riemannian-based Discriminant Analysis (RDA)} %算法的题目 
	\label{alg1} %算法的标签 
	\begin{algorithmic}[1] %此处的[1]控制一下算法中的每句前面都有标号 
		\REQUIRE image dataset $ X\in \mathbb{R}^{D\times N} $, sample labels $ L\in \mathbb{R}^{N\times 1} $ \\
		%\ENSURE  core tensor $ \mathcal{K}\in R^{J_1\times\cdots\times J_{N-1}\times M} $, and projection matrices $ A_n\in R^{I_n\times J_n},\: n=1,\ldots,N-1 $\\
		\STATE initial matrix $ U $, gradient norm tolerance $ \varepsilon ^1=10^{-5} $, and max iteration number $ \mathrm{maxit}=200 $. Let $ 0< c< 1 $, $ \beta ^1=0 $, $ \xi ^0=0 $.
		\FOR{$ k\leq \mathrm{maxit} $} 
		\STATE Compute Hessian in the Euclidean space, \re{$\mathrm{Hess}f\left ( U \right )\left [ \xi  \right ]$}, by Eq.(\ref{cost13}) 
		\STATE Compute the Riemannian Hessian, \re{$\mathrm{hess}f\left ( U \right )\left [ \xi  \right ]$}, by Eq.(\ref{cost14}) 
		\STATE Compute the weighted value \\$ \beta ^k=tr\left ( \eta ^{kT}\eta ^k \right )/tr\left ( \eta ^{(k-1)T}\eta ^{k-1} \right ) $.
		\STATE Compute a transport direction \\$ \mathcal{T}_{U^{k-1}\rightarrow U^k\left ( \xi ^{k-1} \right )}=\mathcal{P}_{U^k}\left ( \xi ^{k-1} \right ) $.
		\STATE Compute a conjugate direction \\$ \xi ^k=-\mathrm{grad}_{\mathrm{R}}f\left ( U^k \right )+\beta ^k\mathcal{T}_{U^{k-1}\rightarrow U^k\left ( \xi ^{k-1} \right )} $.
		\STATE Compute Armijo step size $ \alpha ^k $ using backtracking \\$ f\left ( \mathrm{R}_{U^k}\left ( \alpha ^k\xi ^k \right ) \right )\geq f\left ( U^k \right )+c\alpha ^ktr\left ( \eta ^{kT}\xi ^k \right ) $.
		\STATE Terminate and output $ U^{k+1} $ if one of the stopping conditions, $ \left \| \eta ^{k+1} \right \|_F^2\leq \varepsilon ^1 $, or iteration number $ k\geq \mathrm{maxit} $ is met.
		\ENDFOR
		\STATE OUTPUT $ U $.
	\end{algorithmic} 
\end{algorithm}

\section{Numerical Experiments and Results}
In this section, we test the effectiveness of RDA on feature extraction tasks and classification tasks. RDA is compared with four variants of multilinear discriminant analysis (\textit{i.e.} HODA \cite{phan2010tensor}, DATER \cite{yan2005discriminant}, CMDA \cite{li2014multilinear}, and MHODA \cite{yin2020high}) and four variants of tensor decomposition (\textit{i.e.} NTD \cite{kim2007nonnegative}, LRRHTD \cite{zhang2017low}, HTD Multinomial \cite{sun2015heterogeneous}, and HOSVD \cite{de2000multilinear}).  All subsequent numerical experiments are carried out on a desktop (Intel Core i5-5200U CPU with a frequency of 2.20 GHz and a RAM of 8.00 GB). Each experiment is repeated 10 times, each time using different random sampling data.

\subsection{Datasets Description}
\label{exp:datasets}
Our experiment involves seven benchmark image datasets, namely the COIL20 Object, ETH80 Object, MNIST Digits, USPS Digits, ORL Faces, Olivetti Faces, and CMU PIE Faces. \textbf{Figure 2} shows some examples of sampling from these data sets. We did not show the MNIST dataset here because it is a well-known dataset.

\begin{figure}[htb]
      \centering
      \includegraphics[width=1\linewidth]{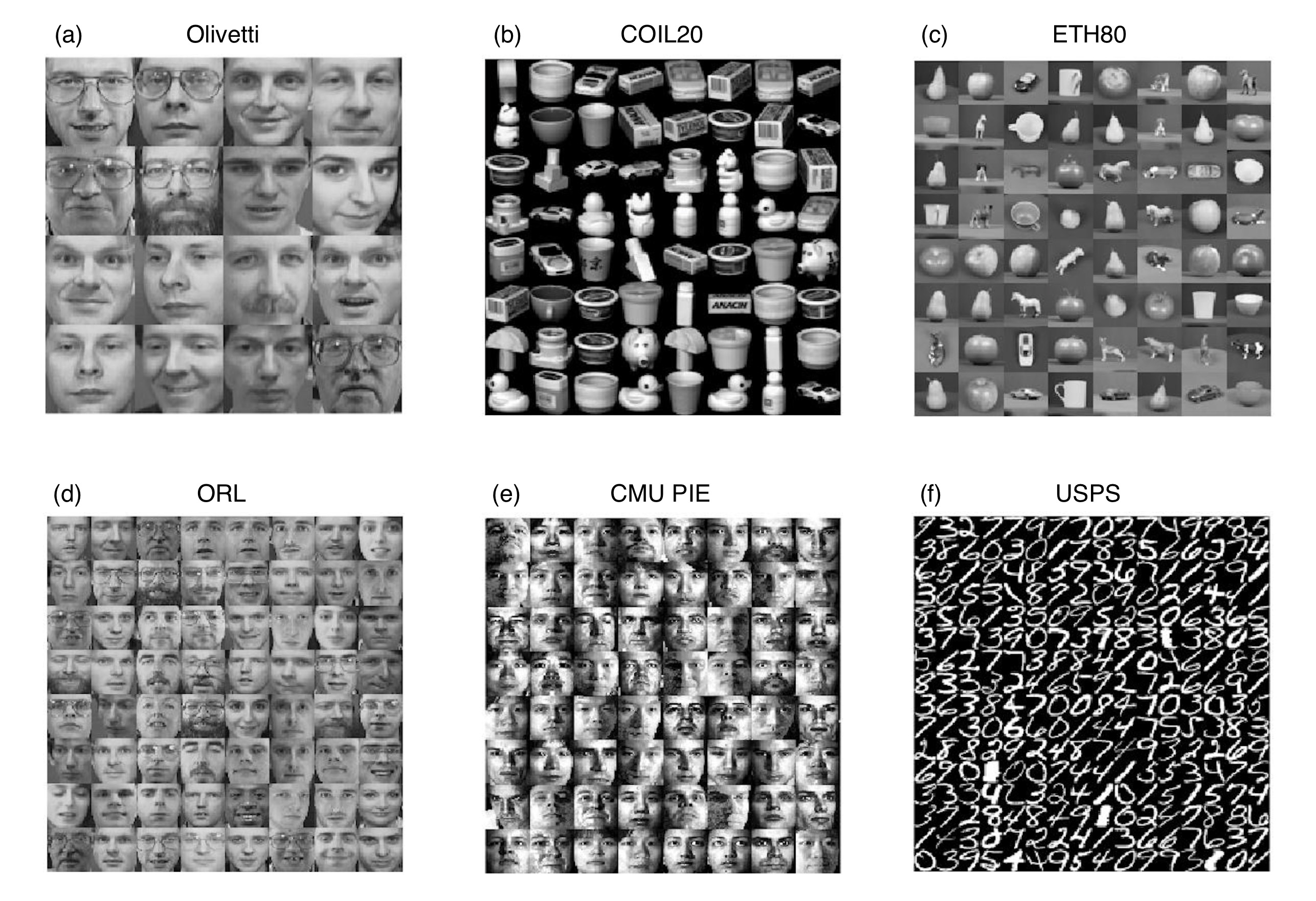}
      \caption{Some examples from six datasets used in experiments. (a) Olivetti dataset. (b) COIL20 dataset. (c) ETH80 dataset. (d) ORL dataset. (e) CMU PIE dataset. (f) USPS dataset.}
      \label{fig:photo}
\end{figure}

The COIL20 dataset contains 1420 grayscale images of 20 objects (72 images per object). Objects in COIL20 have a variety of complex geometric and reflective properties. In our experiments, the image from COIL20 is downsampled to a size of $ 32\times 32 $ with 0-255 grayscale.
 
The ETH80 dataset is a multi-view image dataset used for object classification. It includes 8 categories: apple, car, cow, cup, dog, horse, pear, and tomato. Each category contains 10 objects, and each object has 41 images from different views, resulting in a total of 3280 images. The resolution of original images is $128\times 128 $, and we adjust the size of each image to $32\times 32 $ pixel.

Both USPS and MNIST datasets are 0-9 handwritten digits. The USPS dataset has 11,000 images, with a size of $16\times 16 $ pixels, while the MNIST dataset has 60,000 images belonging to the training set, with  a size of $28\times 28 $ pixels. In our experiment, we randomly selected 2000 images (200 images per category) from the USPS dataset, and 3000 images (300 images per category) from the MNIST dataset.

The ORL dataset contains 400 images \re{from} 40 different people, each with 10 different images. These images were taken multiple times under different lighting conditions and facial expressions (eyes open/closed; with/without smile) and facial details (with/without glasses). All images were taken against a dark uniform background, with the subject in an upright frontal position (tolerance to certain lateral movements). We adjust the size of each image to $ 32\times 32 $ pixels.

The Olivetti dataset consists of 400 faces from 40 people (10 per person). The viewing angle of those images changes very little, but people’s expressions change a lot, and occasionally they wear glasses. The size of the image is $ 64\times 64 =4096 $ pixels, and the data is labeled according to the identity.

The CMU PIE dataset is a gray-scale face dataset, including 68 people, and each person has 141 face images. The images were taken under different lighting conditions. We extracted a subset of 50 individuals and the corresponding 50 facial images of each person, resulting in a total of 2500 images.

% Table 1
\begin{table}
  \caption{Illustrations of the datasets}
  \label{tab:table1}
  \centering
  \small
  \begin{tabular}{ccccc}
    \toprule
   % \multicolumn{3}{c}{Part}                   \\
    \cmidrule(r){1-2}
    dataset     & \#samples  & size$_{original}$   & size$_{final}$  &  \#classes\\
    \midrule
    ETH80       & 3280          & 32*32          & 8*8                  &  8             \\
    MNIST       & 3000          & 28*28          & 10*10                &  10            \\
    USPS        & 2000          & 16*16          & 7*8                  &  10            \\
    COIL20      & 1440          & 32*32          & 8*8                  &  20            \\
	ORL         & 400           & 32*32          & 6*6                  &  40            \\
	Olivetti    & 400           & 64*64          & 8*8                  &  40            \\ 
	CMU PIE     & 2500          & 32*32          & 8*8                  &  50            \\ 
    \bottomrule
  \end{tabular}
\end{table}

\textbf{Table \ref{tab:table1}} shows the general description of seven datasets, where the attributes of each data set are the total number of samples, the dimensions of the original data, the final dimension after dimensionality reduction, and the number of classes we use experiments. Note that each sample has a real category label (such as object, identity, or digit). We preprocess the dataset in the following way: a) randomly shuffle all the data, b) normalize the gray value of pixels to \re{the} unit.

\re{In the following numerical experiments, the data is represented by a third-order tensor, where the first two modes are associated with the spatial information of image pixels, and the last mode represents the number of samples. It is worth noting that RDA algorithm and its implementation are very general, and there is no such restriction on the data format.} In the tests, we first perform subspace learning and reduce the dimensionality of the tensor data (from size$_{original}$ to size$_{final}$ in Table \ref{tab:table1}), and then apply the $k$-means clustering or $k$-nearest-neighbour classification on the extracted low-dimensional features.

\subsection{Clustering analysis}
\label{exp:clustering}
We first test whether the features in the low-dimensional subspace extracted by RDA can cluster the data. Specifically, we use five supervised algorithms (\textit{e.g.} RDA, HODA, CMDA, MHODA, and DATER) to perform subspace learning for each data set. Then we cluster the features on the subspace with $ k\mathrm{-means} $ clustering. We randomly initialize 10 times and calculate the average result of 10 times. The results are quantified by clustering accuracy (ACC) and normalized mutual information (NMI) \cite{sun2015heterogeneous}.

\begin{table*}[htb]
	\centering
	\caption{$ k\mathrm{-means} $ clustering results of RDA and four supervised algorithms on 7 datasets.}~\label{tab:table2}
	\centering \small
	\setlength{\tabcolsep}{6pt}{
		\begin{tabular}{c|c|cc|ccc}
			\hline
			\multirow{2}{*}{Dataset} & \multirow{2}{*}{Metric} & \multicolumn{2}{c|}{Riemannian-based optimization}  & \multicolumn{3}{c}{Euclidean-based optimization} \\ \cline{3-7} 
				          &                   &RDA                  &MHODA           &HODA            &CMDA            &DATER  \\ \hline
			\multirow{2}{*}{ETH80} & ACC      &\bf{0.5452±0.0048}   &0.5098±0.0000   &0.4750±0.0039   &0.4852±0.0108   &0.4714±0.0219  \\ 
			             & NMI                &\bf{0.5094±0.0000}   &0.4691±0.0000   &0.4523±0.0050   &0.4598±0.0102   &0.4155±0.0180  \\ \hline
			\multirow{2}{*}{MNIST} & ACC      &\bf{0.7552±0.0029}   &0.1888±0.1107   &0.5563±0.0297   & *              & *              \\ 
				          & NMI               &\bf{0.6314±0.0016}   &0.0830±0.1256   &0.4902±0.0184   & *              & *              \\ \hline
			\multirow{2}{*}{USPS} & ACC       &\bf{0.8482±0.0010}   &0.5074±0.0673   &0.4580±0.0339   &0.3377±0.0152   &0.4912±0.0570  \\ 
				          & NMI               &\bf{0.7339±0.0000}   &0.4621±0.0718   &0.4368±0.0289   &0.2752±0.0142   &0.4607±0.0447  \\ \hline
			\multirow{2}{*}{COIL20} & ACC     &\bf{0.7948±0.0398}   &0.7244±0.0345   &0.6144±0.0216   &0.6563±0.0324   &0.6337±0.0178  \\ %\cline{2-7} 
				          & NMI               &\bf{0.8553±0.0199}   &0.8133±0.0139   &0.7388±0.0118   &0.7637±0.0093   &0.7334±0.0144  \\ \hline
			\multirow{2}{*}{ORL} & ACC        &\bf{0.7380±0.0278}   &0.5817±0.0262   &0.4437±0.0213   &0.4390±0.0199   &0.4690±0.0273  \\ 
				              & NMI           &\bf{0.8739±0.0112}   &0.7871±0.0114   &0.6769±0.0089   &0.6713±0.0149   &0.6538±0.0194  \\ \hline
			\multirow{2}{*}{Olivetti} & ACC   &\bf{0.7508±0.0407}   &0.6627±0.0372   &0.4900±0.0324   &0.5045±0.0292   &0.5727±0.0404  \\ 
				      & NMI                   &\bf{0.8776±0.0146}   &0.8251±0.0154   &0.7044±0.0152   &0.7155±0.0151   &0.7470±0.0255  \\ \hline
			\multirow{2}{*}{CMU PIE	} & ACC   &\bf{0.7866±0.0220}   &0.5927±0.0193   &0.1546±0.0034   &0.1206±0.0042   &0.3764±0.0299  \\ 
			          & NMI                   &\bf{0.8776±0.0086}   &0.7472±0.0073   &0.3686±0.0078   &0.3014±0.0040   &0.5690±0.0238  \\ \hline
	\end{tabular}}
	\\ \footnotesize{* The algorithm failed in the dataset, as the between-class matrix is singular.}
\end{table*}

\begin{figure}[htb]
      \centering
      \includegraphics[width=1\linewidth]{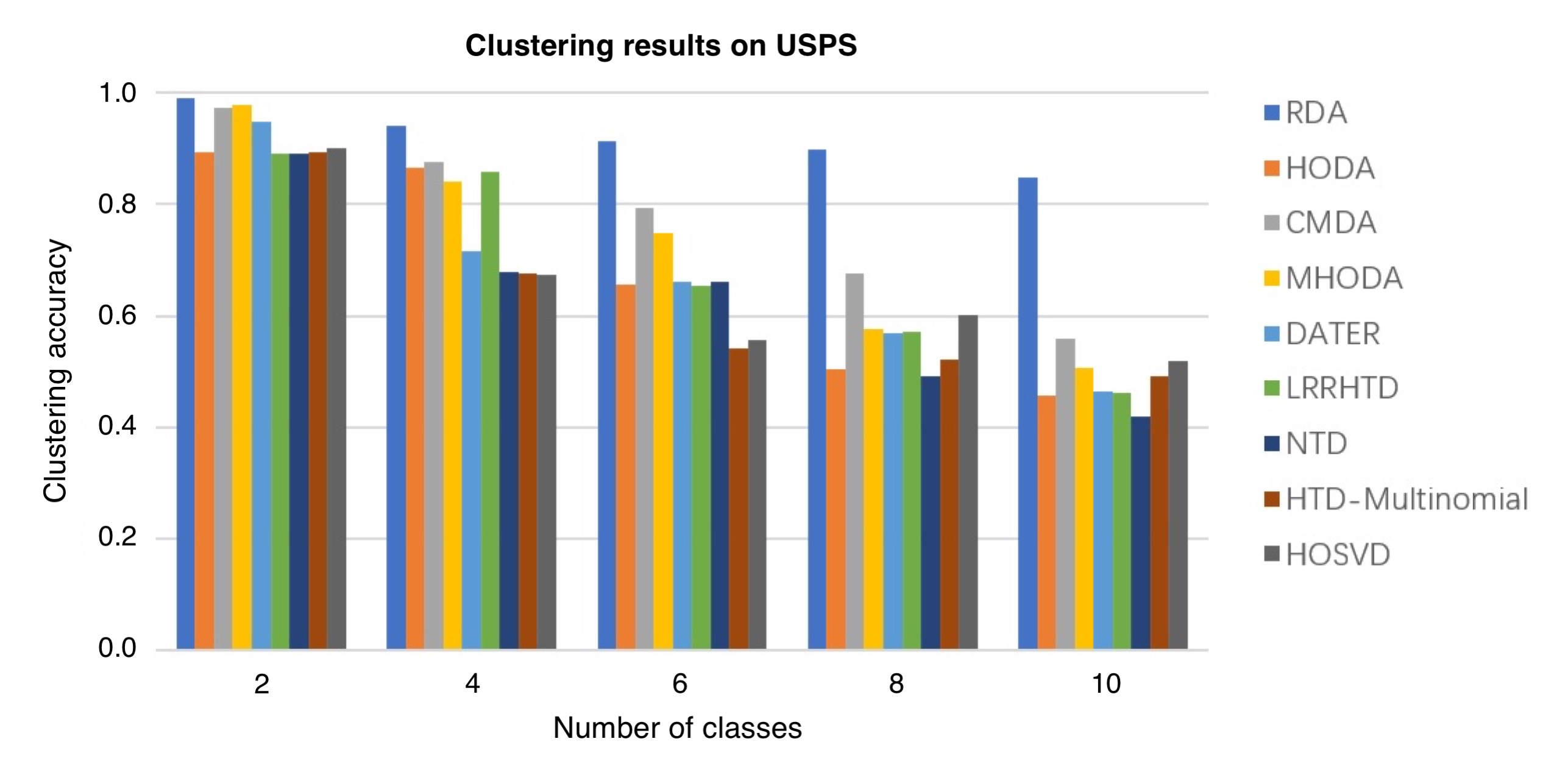}
      \caption{The clustering accuracy varying with the number of classes in the USPS digit dataset. RDA achieves the highest accuracy in digit clustering.}%The results of clustering the USPS data set. We use RDA and eight existing SOTA methods to perform subspace learning on the USPS data set, and then perform $k$-means clustering on the extracted features. RDA has the highest accuracy in digit clustering.}
      \label{fig:USPS}
\end{figure}

\begin{figure}[htb]
      \centering
      \includegraphics[width=1\linewidth]{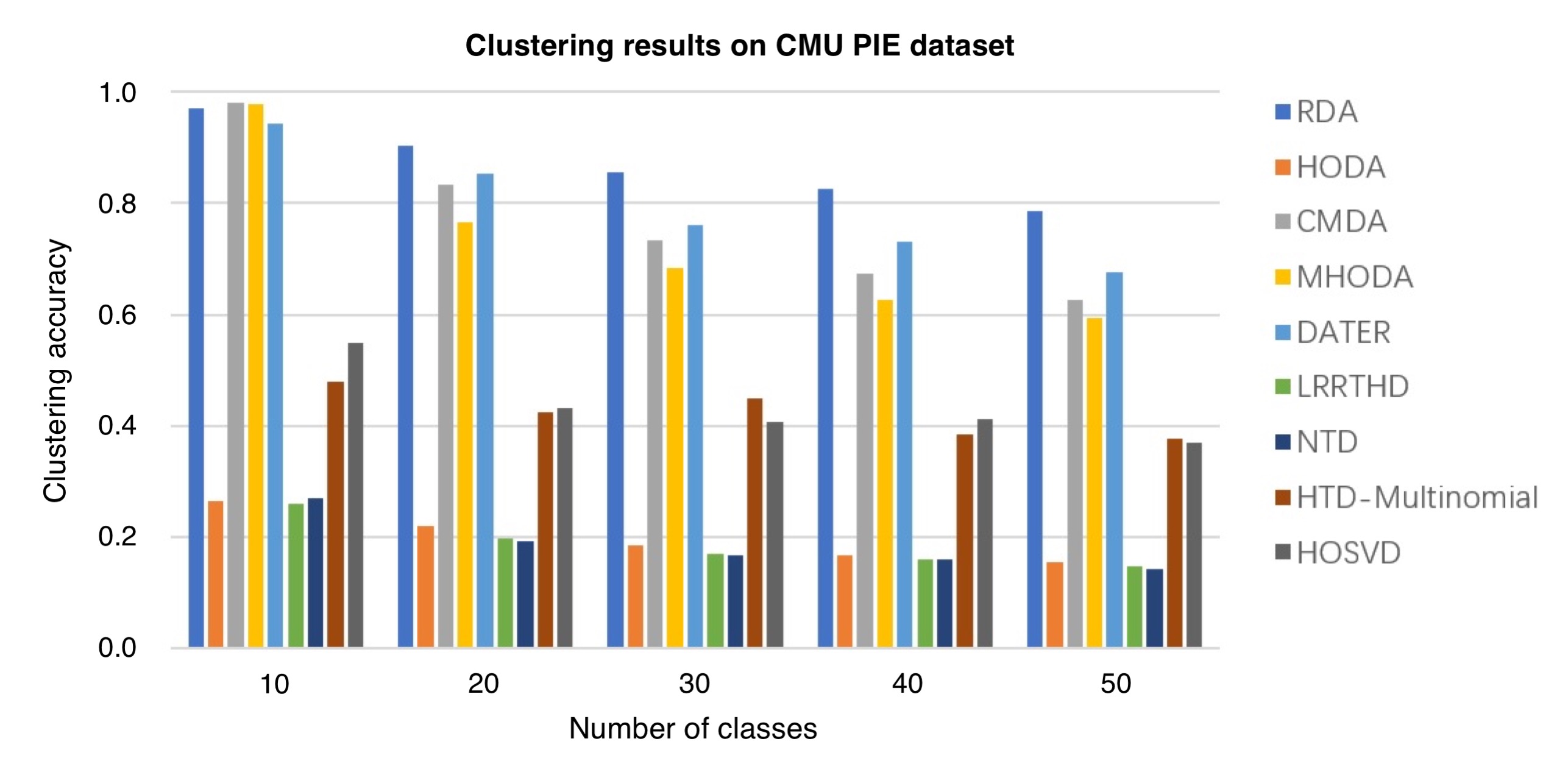}
      \caption{The clustering accuracy varying with the number of classes in the CMU PIE face dataset. RDA achieves the highest accuracy in face clustering.}
      \label{fig:CMU}
\end{figure}

\textbf{Table \ref{tab:table2}} shows the clustering results of RDA and four supervised methods on seven datasets. We show the mean and standard deviation of ACC/NMI in 10 tests. The best result for each data set is highlighted in bold text. Obviously, RDA achieves the best performance compared to HODA, CMDA, MHODA and DATER. Especially, when the dataset is complex and multi-class, such as the CMU PIE dataset, the Riemannian-based algorithms (both of RDA and MHODA) provide better clustering results than Euclidean-based algorithms, implying that Riemannian-based methods have a higher ability to extract complex features. %Therefore, Riemannian-based optimization can obtain an explicit and unique solution. 

We then further test the performance of RDA with four unsupervised tensor decomposition methods, including a Riemannian-based method (\textit{e.g.} HTD-Multinomial), and three Euclidean-based clustering methods (\textit{e.g} LRRHTD, NTD, and HOSVD). \textbf{Table \ref{tab:table3}} shows the experimental results, suggesting that RDA outperforms all the other tested methods.

\begin{table*}[]
	\centering
	\caption{$ k\mathrm{-means} $ clustering results of RDA and four unsupervised tensor decomposition methods on 7 datasets.}~\label{tab:table3}
	\centering \small
	\setlength{\tabcolsep}{5.5pt}{
		\begin{tabular}{c|c|cc|ccc}
			\hline
			\multirow{2}{*}{Dataset} & \multirow{2}{*}{Metric} & \multicolumn{2}{c|}{Riemannian-based optimization}  & \multicolumn{3}{c}{Euclidean-based optimization} \\
			\cline{3-7} 
				          &                   &RDA                  &HTD-Multinomial &LRRHTD          &NTD             &HOSVD          \\ \hline
			
			\multirow{2}{*}{ETH80} & ACC      &\bf{0.5452±0.0048}   &0.4714±0.0219   &0.4994±0.0062   &0.4385±0.0042   &0.4633±0.0025  \\ %\cline{2-7} 
			             & NMI                &\bf{0.5094±0.0000}   &0.4155±0.0180   &0.4764±0.0065   &0.3968±0000     &0.3773±0000    \\ \hline
		\multirow{2}{*}{MNIST} & ACC      &\bf{0.7552±0.0029}   &0.5040±0.0385   &0.5365±0.0135   &0.5090±0.0140   &0.5101±0.0023  \\ %\cline{2-7} 
				          & NMI               &\bf{0.6314±0.0016}   &0.4386±0.0247   &0.4790±0.0054   &0.4608±0.0053   &0.4484±0.0024  \\ \hline
		\multirow{2}{*}{USPS} & ACC       &\bf{0.8482±0.0010}   &0.4912±0.0570   &0.4625±0.0089   &0.4186±0.0311   &0.5200±0.0259  \\ %\cline{2-7} 
				          & NMI               &\bf{0.7339±0.0000}   &0.4607±0.0447   &0.4699±0.0064   &0.4324±0.0199   &0.4639±0.0142  \\ \hline
\multirow{2}{*}{COIL20} & ACC     &\bf{0.7948±0.0398}   &0.6337±0.0178   &0.6633±0.0296   &0.6317±0.0265   &0.5928±0.0199  \\ %\cline{2-7} 
				          & NMI               &\bf{0.8553±0.0199}   &0.7334±0.0144   &0.7675±0.0116   &0.7428±0.0122   &0.7215±0.0153  \\ \hline	
			\multirow{2}{*}{ORL} & ACC        &\bf{0.7380±0.0278}   &0.4690±0.0273   &0.5215±0.0252   &0.4397±0.0186   &0.5915±0.0284  \\ %\cline{2-7} 
				              & NMI           &\bf{0.8739±0.0112}   &0.6538±0.0194   &0.7339±0.0127   &0.6704±0.0112   &0.7611±0.0239  \\ \hline
			\multirow{2}{*}{Olivetti} & ACC   &\bf{0.7508±0.0407}   &0.5727±0.0404   &0.5300±0.0309   &0.5627±0.0163   &0.5693±0.0266  \\ %\cline{2-7} 
				      & NMI                   &\bf{0.8776±0.0146}   &0.7470±0.0255   &0.7347±0.0166   &0.7366±0.0092   &0.7451±0.0156  \\ \hline
			\multirow{2}{*}{CMU PIE} & ACC    &\bf{0.7866±0.0220}   &0.3764±0.0299   &0.1477±0.0041   &0.1424±0.0025   &0.3707±0.0277  \\ %\cline{2-7} 
				          & NMI               &\bf{0.8776±0.0086}   &0.5690±0.0238   &0.3521±0.0063   &0.3420±0.0040   &0.5994±0.0163  \\ \hline
	\end{tabular}}
\end{table*}

We further investigate the influence of the number of classes on the performance of RDA clustering. We test the clustering ability of RDA on the USPS digit dataset and CUM PIE face dataset, compared with other seven SOTA algorithms. \textbf{Figure~\ref{fig:USPS}-\ref{fig:CMU}} shows that the clustering accuracy varying with the number of classes in the USPS dataset and CMU PIE dataset, respectively. These results confirm that RDA robustly achieves the best performance on both data sets regardless of the number of classes.

%Due to the full utilization of sample labels and the discovery of non-linear structures by RDA, our proposed RDA is superior to traditional methods optimized in Euclidean space (such as HODA, CMDA, DATER in Table 2). In addition, the existing Riemannian-based algorithms (\textit{e.g.} HTD-Multinomial and MHODA), as well as the tensor decomposition methods (\textit{e.g.} LRRHTD and NTD), are not as good as RDA algorithms. It is worth noting that RDA obtains higher performance for dealing with multi-class and complex dataset (\textit{e.g.} CMU PIE, COIL20).
 
\subsection{Classification}
\label{exp:classification}
%\wg{Please check Quanying's revision. Thank you.}
\re{Here we test the classification performance using the learned features from RDA with a standard classifier, namely $ k $-nearest-neighbour ($ k\mathrm{NN} $) classifier. } We calculate the projection matrix $ U $ from the train samples $ X_{train} $, and then use the learned matrix $ U $ to learn the low-dimensional representation of the test data $ X_{test} $. The class of the test samples is predicted with the following equation: 

\begin{equation}\label{cost19}
Y_{test}=U^TX_{test}
\end{equation}

We conduct classification experiments on five benchmark datasets (\textbf{Sec}~\ref{exp:datasets}), including ETH80, MNIST, USPS, COIL 20 and CMU PIE. \re{The data samples from the datasets are assumed to have the uniform distribution in each experiment.} A 3-fold cross validation is applied to the training data and a 5-fold cross validation to the test data. We use the the ACC, NMI and $ k\mathrm{NN} $ classification accuracy as the evaluation metrics. 

\textbf{Table}~\ref{tab:table4} shows the classification results from RDA and other methods \re{using a $ k\mathrm{NN} $ classifier}. As shown in \textbf{Table}~\ref{tab:table4}, RDA achieves better performance than most existing algorithms. Interestingly, the MHODA algorithm, optimizing via the product manifold, is consistently worse than RDA on the Stiefel manifold, implying that the Stiefel manifold optimization might be more robust than the product manifold. 

% \usepackage{multirow}
\begin{table*}[]
	\centering
	\caption{Comparisons of classification results on 5 datasets.}~\label{tab:table4}
		\centering \small
	\setlength{\tabcolsep}{5.5pt}{
	\begin{tabular}{c|c|cc|cccc}
		\hline
		\multirow{2}{*}{Dataset} & \multirow{2}{*}{Metric} & \multicolumn{2}{c|}{\footnotesize Riemannian-based optimization}  & \multicolumn{4}{c}{\footnotesize Euclidean-based optimization} \\ \cline{3-8} 
        	          &                          &RDA         &MHODA   &HODA    &CMDA   &DATER       &HOSVD        \\ \hline
		
		\multirow{3}{*}{ETH80} &ACC              &\bf{0.5405} &0.5058  &0.4784  &0.5170 &0.5104      &0.4665       \\ %\cline{2-8} 
		             &NMI                        &\bf{0.5073} &0.4692  &0.4489  &0.4565 &0.4571      &0.3816       \\ %\cline{2-8} 
		                  &$k\mathrm{NN}$        &0.7355      &0.6856  &0.7621  &0.7650 &0.7686      &\bf{0.7844}  \\ \hline
		\multirow{3}{*}{MNIST} &ACC              &\bf{0.7631} &0.2641  &0.5494  &*      &*           &0.5114       \\ %\cline{2-8} 
		             &NMI                        &\bf{0.6509} &0.1700  &0.4875  &*      &*           &0.4565       \\ %\cline{2-8} 
		                  &$k\mathrm{NN}$        &0.8445      &0.4040  &0.8505  &*      &*           &\bf{0.8555}  \\ \hline
		\multirow{3}{*}{USPS} &ACC               &\bf{0.8600} &0.5266  &0.4554  &0.5688 &0.5487      &0.5026       \\ %\cline{2-8} 
		              &NMI                       &\bf{0.7565} &0.4889  &0.4363  &0.5352 &0.5285      &0.4682       \\ %\cline{2-8} 
		                  &$k\mathrm{NN}$        &0.8591      &0.5952  &0.7878  &0.8808 &\bf{0.8831} &0.8458       \\ \hline
		\multirow{3}{*}{COIL20} &ACC             &\bf{0.7777} &0.6973  &0.6155  &0.7247 &0.7442      &0.6050       \\ %\cline{2-8} 
		            &NMI                         &\bf{0.8522} &0.8385  &0.7402  &0.8264 &0.8378      &0.7163       \\ %\cline{2-8} 
		                  &$k\mathrm{NN}$        &\bf{0.8771} &0.8385  &0.6729  &0.8417 &0.8302      &0.7177       \\ \hline
		\multirow{3}{*}{CMU PIE} &ACC            &\bf{0.8024} &0.5866  &0.1708  &0.6166 &0.6018      &0.3920       \\ %\cline{2-8} 
		           &NMI                          &\bf{0.8857} &0.7435  &0.3848  &0.7638 &0.7625      &0.6061       \\ %\cline{2-8} 
		                  &$k\mathrm{NN}$        &\bf{0.6713} &0.5261  &0.2147  &0.6002 &0.6205      &0.1890       \\ \hline
	\end{tabular}}
	\\ \footnotesize{* The algorithm failed in the dataset, as the between-class matrix is singular.}
\end{table*}

To compare the first-order approximation and the second-order approximation, we test the trust region methods (RDA and MHODA) and the conjugate gradient methods (conj-RDA and conj-MHODA). As shown in \textbf{Table \ref{tab:first-vs-second}}. We find that RDA reliably outperforms conj-RDA in all datasets, suggesting the trust region method is better than the first-order approximation for RDA. The improvement of the trust region method is not obvious for the manifold-based high-order discriminant analysis (\textit{i.e} MHODA vs conj-MHODA), which might be caused by the sub-optimal solution in the product manifold optimization used in MHODA.

% \usepackage{multirow}
\begin{table*}[]
	\centering
	\caption{Comparison of algorithmic performance on first-order approximation and the second-order approximation.}~\label{tab:first-vs-second}
		\centering \small
	\setlength{\tabcolsep}{5.5pt}{
	\begin{tabular}{c|c|cc|cc}
		\hline
		\multirow{2}{*}{Dataset} & \multirow{2}{*}{Metric} & \multicolumn{2}{c|}{Stiefel manifold optimization}  & \multicolumn{2}{c}{Product manifold optimization} \\ \cline{3-6} 
        	          &          &RDA                 &conj-RDA       &MHODA           &conj-MHODA        \\ \hline
		\multirow{3}{*}{ETH80} &ACC             &\bf{0.5000±0.0200}  &0.4362±0.0284  &0.5099±0.0091   &0.5012±0.0086     \\ %\cline{2-6} 
		             &NMI             &\bf{0.5231±0.0050}  &0.4749±0.0197  &0.4696±0.0059   &0.4674±0.0052     \\ %\cline{2-6} 
		                  &$k\mathrm{NN}$  &\bf{0.7156±0.0226}  &0.6935±0.0435  &0.6699±0.0294   &0.6747±0.0322     \\ \hline
		\multirow{3}{*}{MNIST} &ACC             &\bf{0.7696±0.0207}  &0.6559±0.0343  &0.2148±0.1331   &0.1679±0.0383     \\ %\cline{2-6} 
		             &NMI             &\bf{0.6518±0.0145}  &0.5634±0.0160  &0.1137±0.1431   &0.0726±0.0617     \\ %\cline{2-6} 
		                  &$k\mathrm{NN}$  &\bf{0.8420±0.0236}  &0.8110±0.0244  &0.4410±0.1116   &0.3620±0.0461     \\ \hline
		\multirow{3}{*}{USPS} &ACC             &\bf{0.8599±0.0062}  &0.6931±0.0609  &0.4978±0.0816   &0.5162±0.0249     \\ %\cline{2-6} 
		              &NMI             &\bf{0.7555±0.0053}  &0.6097±0.0378  &0.4682±0.0764   &0.4888±0.0057     \\ %\cline{2-6} 
		                  &$k\mathrm{NN}$  &\bf{0.8724±0.0414}  &0.8192±0.0380  &0.6091±0.0516   &0.6013±0.0411     \\ \hline
		\multirow{3}{*}{COIL20} &ACC             &\bf{0.7666±0.0376}  &0.6473±0.0373  &0.7111±0.0495   &0.7085±0.0291     \\ %\cline{2-6} 
		            &NMI             &\bf{0.8469±0.0177}  &0.7506±0.0204  &0.8133±0.0191   &0.8077±0.0127     \\ %\cline{2-6} 
		                  &$k\mathrm{NN}$  &\bf{0.8625±0.0380}  &0.7438±0.0247  &0.8385±0.0445   &0.8229±0.0396     \\ \hline
		\multirow{3}{*}{CMU PIE} &ACC             &\bf{0.8543±0.0306}  &0.3526±0.0436  &0.5784±0.0233   &0.5934±0.0140     \\ %\cline{2-6} 
		         &NMI             &\bf{0.9344±0.0100}  &0.5767±0.0343  &0.7414±0.0128   &0.7485±0.0095     \\ %\cline{2-6} 
		                  &$k\mathrm{NN}$  &\bf{0.7973±0.0422}  &0.3923±0.0445  &0.5512±0.0522   &0.5659±0.0333     \\ \hline
	\end{tabular}}
\end{table*}

\subsection{Sparse regularized RDA}
The sparsity property has been reported in many real-world applications, and using sparsity regularization have the advantages of being robust to noise and thus might improve the classification performance especially for the high-dimensional data \cite{chen2018solving}. In order to study the effect of sparsity regularization on RDA-based classification, we apply the second-order geometry of the trust-region method and the first-order geometry of the conjugate gradient to solve the loss function in Eq. (\ref{cost15}) on Stiefel manifold and Grassmann manifold, respectively. \textbf{Table \ref{tab:sparse}} lists the classification performance of the sparsity regularized RDA. StRDA and GrRDA represent to the RDA on Stiefel manifold and Grassmann manifold, respectively. SStRDA and SGrRDA represent StRDA and GrRDA with an additional sparsity regularization, while conj-SStRDA and conj-SGrRDA denote SStRDA and SGrRDA solved by the first-order geometry of the conjugate gradient method.

In theory, sparsity regularization on $U$ can reduce the learning parameters and improve the generalization ability of algorithms \cite{chen2018solving}. Our experimental evidence in (\textbf{Table \ref{tab:first-vs-second} \& \ref{tab:sparse}}) also supports the sparsity regularization of Stiefel manifold (StRDA vs SStRDA) and Grassmann manifold (GrRSA vs SGrRDA) in most cases,  demonstrating that the sparsity regularization can effectively enhance model generalization, which is consistent with previous study \cite{chen2018solving}.

% \usepackage{multirow}
\begin{table*}[]
	\centering
	\caption{Comparison of classification results \re{with/without a sparsity regularization term.}}~\label{tab:sparse}
		\centering \small
	\setlength{\tabcolsep}{5.5pt}{
	\begin{tabular}{c|c|ccc|ccc}
		\hline
		\multirow{2}{*}{Dataset} & \multirow{2}{*}{Metric} & \multicolumn{3}{c|}{Stiefel manifold Optimization} & \multicolumn{3}{c}{Grassman manifold Optimization}\\ \cline{3-8} 
        	          &          &StRDA           &SStRDA        &conj-SStRDA   &GrRDA        &SGrRDA         &conj-SGrRDA       \\ \hline
		\multirow{3}{*}{COIL20} &ACC             &0.7666          &0.7851        &0.7721        &0.7818       &\bf{0.7872}    &0.7713         \\ %\cline{2-8} 
		            &NMI             &0.8469          &0.8556        &0.8488        &0.8523       &\bf{0.8573}    &0.8512         \\ %\cline{2-8} 
		                  &$k\mathrm{NN}$  &0.8625          &0.8562        &0.8615        &0.8562       &\bf{0.8844}    &0.8406         \\ \hline
		\multirow{3}{*}{ETH80} &ACC             &0.5000          &0.4959        &0.4970        &\bf{0.5029}  &0.4935         &0.4970         \\ %\cline{2-8} 
		             &NMI             &0.5231          &0.5260        &0.5218        &\bf{0.5260}  &0.5253         &0.5240         \\ %\cline{2-8} 
		                  &$k\mathrm{NN}$  &0.7156          &0.7194        &0.7156        &\bf{0.7301}  &0.7226         &0.7171         \\ \hline
		\multirow{3}{*}{Dense ETH80} &ACC             &0.5486          &0.5439        &0.5333        &\bf{0.5489}  &0.5451         &0.5386         \\ %\cline{2-8} 
		       &NMI             &0.5113          &0.5095        &0.5018        &0.5105       &\bf{0.5131}    &0.5059         \\ %\cline{2-8} 
		                  &$k\mathrm{NN}$  &0.7338          &0.7162        &0.6920        &0.7256       &\bf{0.7350}    &0.6674         \\ \hline
		\multirow{3}{*}{MNIST} &ACC             &\bf{0.7696}     &0.7601        &0.7695        &0.7682       &0.7570         &0.7651         \\ %\cline{2-8} 
		             &NMI             &0.6518          &0.6510        &0.6505        &\bf{0.6532}  &0.6465         &0.6491         \\ %\cline{2-8} 
		                  &$k\mathrm{NN}$  &0.8420          &0.8480        &0.8395        &0.8415       &0.8465         &\bf{0.8715}    \\ \hline
		\multirow{3}{*}{USPS} &ACC             &0.8599          &0.8375        &0.8413        &0.8498       &0.8615         &\bf{0.8655}    \\ %\cline{2-8} 
		              &NMI             &0.7555          &0.7434        &0.7492        &0.7546       &0.7578         &\bf{0.7609}    \\ %\cline{2-8} 
		                  &$k\mathrm{NN}$  &0.8724          &\bf{0.8872}   &0.8716        &0.8680       &0.8777         &0.8715         \\ \hline
		\multirow{3}{*}{CMU PIE } &ACC             &0.8543          &\bf{0.8619}   &0.8482        &0.8602       &0.8587         &0.8407         \\ %\cline{2-8} 
		          &NMI             &0.9344          &0.9324        &0.9310        &0.9361       &\bf{0.9381}    &0.9290         \\ %\cline{2-8} 
		                  &$k\mathrm{NN}$  &0.7973          &0.7954        &0.7761        &0.7813       &\bf{0.8328}    &0.8095         \\ \hline
	\end{tabular}}
\end{table*}

\section{Discussion and Conclusion}
In this paper, we proposed a novel method using Riemannian manifold optimization for discriminant subspace learning, namely Riemannian-based discriminant analysis. The numerical results (\textbf{Table 2-4}) suggest that RDA outperforms many other methods optimized in Euclidean space, as well as the existing Riemannian-based methods. 

Since the inter-class scatter matrix may be singularity (such as the CMDA and DATER algorithms shown in \textbf{Table 2\&4}), many traditional Euclidean-based methods for subspace learning may not be able to guarantee monotonic convergence to its optimal solution. The previous literature also reported similar results \cite{sifaou2020high}.  In contrast, our proposed RDA can effectively avoid the singularity problem. As RDA has the subtraction form of the loss function instead of a division form in the traditional methods, RDA can effectively avoid calculating the inverse of Hessian matrix, thereby reducing the amount of calculation to the Riemannian Hessian. 

Due to the discovery of non-linear structures, our proposed RDA is superior to traditional methods optimized in Euclidean space (such as HODA, CMDA, DATER in \textbf{Table}~\ref{tab:table2}). In addition, the existing Riemannian-based algorithms (\textit{e.g.} HTD-Multinomial and MHODA), as well as the tensor decomposition methods (\textit{e.g.} LRRHTD and NTD), are not as good as RDA algorithms. It is worth noting that RDA obtains higher performance for dealing with multi-class and complex dataset (\textit{e.g.} CMU PIE, COIL20). RDA can provide the higher clustering accuracy regardless of the number of classes (\textbf{Figure~\ref{fig:USPS}-\ref{fig:CMU}}), suggesting that Riemannian-based optimization reliably learns an optimal subspace. Generally, the supervised learning methods are superior to unsupervised learning methods in extracting and selecting features (\textbf{Table}~\ref{tab:table2} vs \textbf{Table} \ref{tab:table3}) due to the full utilization of sample labels in the supervised learning, which is in line with previous study~\cite{xu2009semi}. 

The comparisons between the trust-region methods (RDA and MHODA) and the conjugate gradient methods (conjRDA and conj-MHODA) shows that the second-order geometry of the trust-region method improves the clustering and classification performance of RDA, although it may be unreliable for MHODA (\textbf{Table~\ref{tab:first-vs-second}}). The advantage of RDA may stem from the use of equivalence classes in vertical space, which can effectively isolate the optimal solution in the quotient space \cite{absil2009optimization}. 

Although we have shown that RDA benefits from the use of Riemannian geometry in subspace learning, there are many aspects that have not been covered in this study. For example, how to initialize the factor matrix to ensure the convergence of the algorithm, how to design regularization terms other than sparsity, and how to choose the best Riemann metric. It has been shown that adjusting the Riemann metric according to the underlying structure and constraints of the loss function can speed up the convergence speed and reduce the running time \cite{kasai2016low}. Moreover, Riemannian-based discriminant analysis has a limitation: it suffers from an expensive optimization process to find the optimal subspace. In order to solve this problem, other methods, such as Riemannian preconditioning \cite{kasai2016low}, are worthy of further study.

Beyond our work, many other traditional methods solved in the Euclidean space can be transformed into the Riemannian manifold space and employ Riemannian manifold optimization. It is of particular interests to study how to design the cost function and achieve super-linear convergence in the tangent space of each iteration. When designing algorithms in Riemannian manifold, it is important to balance the trade-off between effectiveness (\textit{e.g.} accuracy) and efficiency (\textit{e.g.} computational complexity).

In summary, RDA provide a novel way to solve the LDA problem with Riemannian manifold optimization. It is an effective method for dimensionality reduction, feature extraction, and classification. We believe that RDA for subspace learning method has great potential in many practical applications. %One potential application of RDA is a brain-computer interface based on EEG.

\section*{Acknowledgements}
The authors would like to thank anonymous reviewers for their detailed and helpful comments. This work was funded in part by the National Natural Science Foundation of China (62001205), Guangdong Natural Science Foundation Joint Fund (2019A1515111038), Shenzhen Science and Technology Innovation Committee (20200925155957004, KCXFZ2020122117340001, SGDX2020110309280100), Shenzhen Key Laboratory of Smart Healthcare Engineering (ZDSYS20200811144003009).
%\label{Reference}


\bibliographystyle{model1-num-names}
\bibliography{RDA-ref.bib}






% \end{thebibliography}

\end{document}